\documentclass[11pt]{article}

\usepackage[final]{acl}

\usepackage{times}
\usepackage{latexsym}
\usepackage[T1]{fontenc}
\usepackage[utf8]{inputenc}
\usepackage{microtype}
\usepackage{inconsolata}

\usepackage{graphicx}
\usepackage{subcaption}
\usepackage{booktabs}
\usepackage{url}
\usepackage[table]{xcolor}
\usepackage{bbm}
\usepackage{multirow}
\usepackage[normalem]{ulem}
\usepackage{tabularx}
\usepackage{textcomp}
\usepackage{amsmath}
\usepackage{amssymb}
\usepackage{mathtools}
\usepackage{amsthm}
\usepackage[most]{tcolorbox}
\usepackage{fvextra}
\usepackage{algorithm}
\usepackage{algorithmic}
\usepackage[capitalize,noabbrev]{cleveref}

\definecolor{mygreen}{RGB}{0,128,0}
\definecolor{myred}{RGB}{180,0,0}
\definecolor{lightyellowbg}{rgb}{1.0, 1.0, 0.70}
\definecolor{highlightblue}{RGB}{218, 236, 252}
\definecolor{bg_green}{RGB}{226, 240, 217}
\definecolor{bg_blue}{RGB}{218, 232, 252}
\definecolor{bg_pink}{RGB}{255, 240, 245}

\theoremstyle{plain}

\theoremstyle{definition}

\theoremstyle{remark}

\title{ICA: Information-Aware Credit Assignment for Visually Grounded Long-Horizon Information-Seeking Agents}

\author{
\textbf{Cong Pang\textsuperscript{1,4,*}},
\textbf{Yujie Yi\textsuperscript{3,4,*}},
\textbf{Xuyu Feng\textsuperscript{2,4,*}},
\textbf{Jiaqi Su\textsuperscript{3,4,*}},
\textbf{Zixuan Chen\textsuperscript{4}},
\textbf{Jiawei Hong\textsuperscript{4}},
\\
\textbf{Tiankuo Yao\textsuperscript{4}},
\textbf{Nang Yuan\textsuperscript{3,4}},
\textbf{Jiapeng Luo\textsuperscript{4}},
\textbf{Lewei Lu\textsuperscript{4,\textdagger}},
\textbf{Xin Lou\textsuperscript{1,\textdagger}}
\\
\textsuperscript{1}ShanghaiTech University,
\quad
\textsuperscript{2}Wuhan University,
\\
\textsuperscript{3}Shanghai Jiao Tong University,
\quad
\textsuperscript{4}SenseTime Research
}

\begin{document}
\maketitle
\begingroup
\renewcommand{\thefootnote}{\fnsymbol{footnote}}
\footnotetext[1]{Equal contribution.}
\footnotetext[2]{Corresponding author.}
\endgroup

\begin{abstract}
% Multi-turn reinforcement learning has significantly advanced the performance of information-seeking agents, while two compounding challenges persist as critical bottlenecks: the fidelity loss inherent in text-derived webpage representations and sparse long-horizon credit assignment. To tackle these issues, we introduce a visual-native search framework that represents webpages as visual snapshots, yielding stable, identifiable evidence units that can be reliably compared across trajectories. Building on this consistent representation, we propose Information-Aware Credit Assignment (ICA), a post-hoc algorithm that estimates the utility of acquired evidence units via posterior success association and propagates dense, turn-level rewards back to key retrieval decisions. Integrated with GSPO, our approach consistently outperforms text-based and outcome-only baselines on BrowseComp, GAIA, Xbench-DS, and Seal-0; notably, our 8B model surpasses all open-source agents under 15B on every evaluated benchmark, demonstrating that information-preserving observations and information-aware credit assignment together provide an effective path toward reinforcement learning for long-horizon web agents. Code and datasets will be released at \url{https://github.com/pc-inno/ICA_MM_deepsearch}.
Long-horizon reinforcement learning for information seeking agents remains difficult because terminal rewards reveal whether the final answer is correct, but not which acquired information enabled it. This difficulty is amplified by text-derived webpage observations, where parsing, truncation, and summarization often produce incomplete and unstable content representations across trajectories. We propose an evidence-centric framework for web agent learning that represents information acquired through tools as identifiable units for comparison across trajectories. In particular, fetched webpages are represented as rendered snapshots, preserving layout and multimodal content as stable content-level observations. Building on these units, we introduce Information-Aware Credit Assignment, a post hoc reward propagation method that estimates turn-level utility scores from rollout success rates and assigns dense rewards to intermediate steps that introduced high-utility information. Integrated with GSPO, our method consistently improves performance on BrowseComp, GAIA, Xbench-DS, and Seal-0. Code and datasets will be released at \url{https://github.com/pc-inno/ICA_MM_deepsearch}.
\end{abstract}
\begin{figure*}[t]
    \centering
    \includegraphics[width=0.99\linewidth]{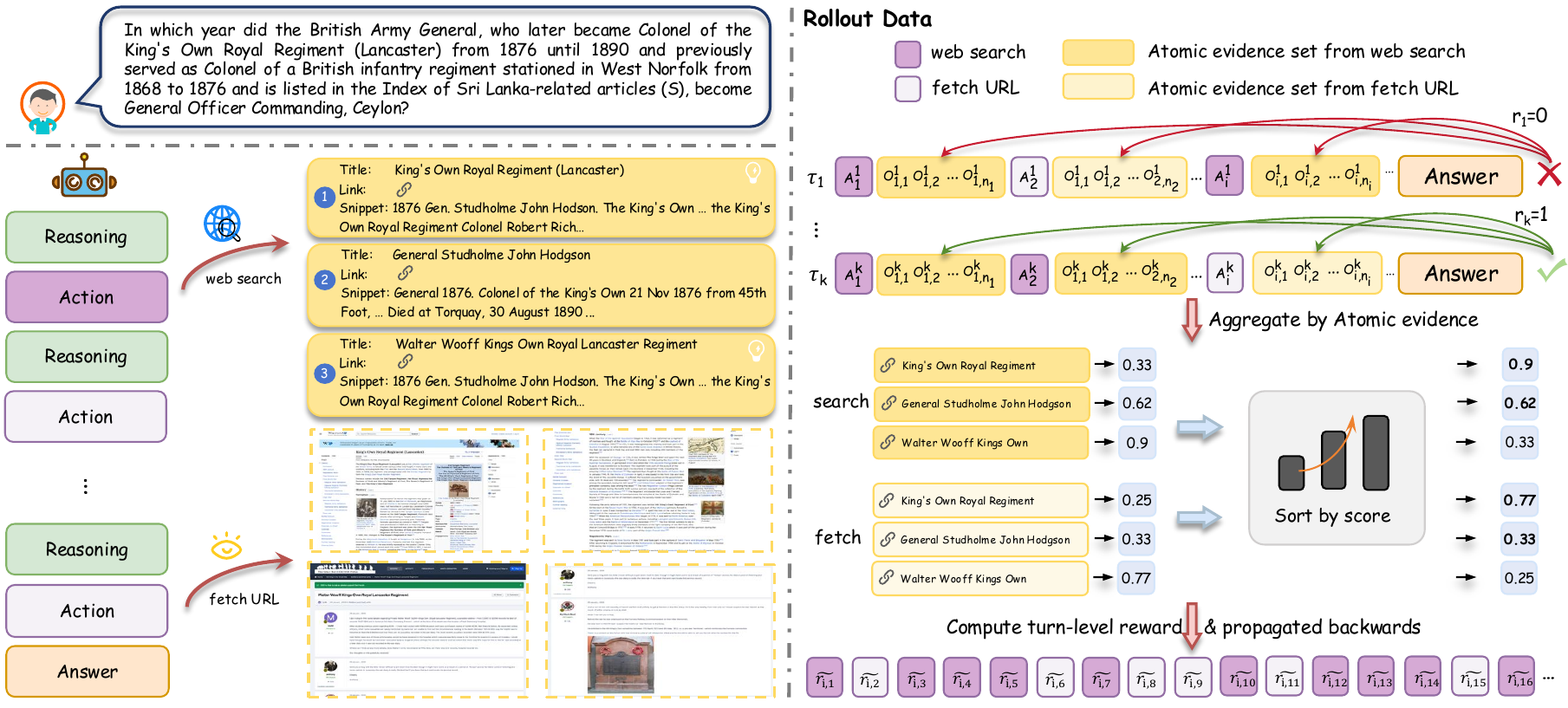}
    \caption{\textbf{ICA Framework.} On the left, an information-seeking agent alternates Reasoning-Action turns, invoking web search and fetch URL to acquire external evidence before producing a final answer. Given a batch of rollout trajectories $\{\tau_i\}$ with sparse outcome supervision, we group interactions by the webpage snapshots obtained from each visited website. We then estimate the marginal utility of each acquired webpage content by its association with successful outcomes, and propagate this signal back to the turns that revealed the content, yielding dense turn-level rewards $r_{i,t}$ for long-horizon credit assignment.}
    \label{fig:pipeline}
\end{figure*}

\section{Introduction}
% Recent systems built on large language models have advanced open-web information seeking toward increasingly sophisticated interactions with external web environments~\cite{he2025webseer, tao2025webleaper, webdancer, webwalker, zhang2026chaining}, demonstrating strong performance on complex multi-hop question answering benchmarks~\cite{kwiatkowski2019natural, press2023measuring, mialon2023gaia, wei2025browsecomp, chen2025xbench, pham2025sealqa} that require iterative retrieval and reasoning. Despite these advances, scaling such agents to long-horizon trajectories remains brittle due to two compounding bottlenecks in observation and optimization. On the observation front, most web agents~\cite{he2025webseer, tongyi2025deepresearch, webdancer, webwalker} rely on text-based parsing pipelines that linearize visually rich webpages into unstructured sequences, discarding layout semantics and introducing representational inconsistencies across trajectories that distort evidence acquisition. On the optimization front, learning is typically guided only by sparse, terminal outcome rewards, which provide little credit assignment to the specific intermediate retrieval decisions that actually contribute relevant evidence amid distractors.
Recent systems built on large language models have advanced open-web information seeking toward increasingly sophisticated interactions with external web environments \citep{he2026webseer,tao2025webleaper,wu2026webdancer,wu-etal-2025-webwalker,zhang2026chainingevidencerobustreinforcement}, demonstrating strong performance on complex multi-hop question answering benchmarks \citep{kwiatkowski-etal-2019-natural,press2023measuring,mialon2024gaia,wei2025browsecomp,chen2025xbenchtrackingagentsproductivity,pham2026sealqaraisingbarreasoning}. However, scaling such agents to long-horizon search remains difficult. In information seeking tasks, intermediate tool actions are useful mainly because they acquire external information that may later support the final answer. Terminal rewards, however, only indicate whether the final answer is correct, leaving unclear which search or fetch decisions introduced information that was useful for successful answer generation. We therefore view long-horizon web agent learning as an evidence attribution problem, where effective optimization requires not only rewarding successful final answers, but also assigning credit to useful retrieval actions along the trajectory. To address this issue, we introduce Information-Aware Credit Assignment (ICA), which assigns credit according to the utility of the external information acquired by each tool action. Rather than treating all intermediate turns as uniformly responsible for the final outcome, ICA estimates which evidence units are associated with successful rollouts and propagates reward to the search and fetch turns that introduced them.

A key prerequisite for ICA is that acquired evidence must be identifiable and comparable across trajectories. However, existing web agents commonly rely on text-based parsing pipelines that convert webpages into linearized text sequences \citep{he2026webseer,tongyideepresearchteam2026tongyideepresearchtechnicalreport,wu2026webdancer,wu-etal-2025-webwalker}. Although convenient for language models, such representations often discard layout structure, visual cues, tables, and other presentation dependent information. They also make evidence identity unstable, since the same webpage may be parsed, truncated, chunked, or summarized differently across tool pipelines and context budgets. We therefore represent fetched webpages as rendered snapshots, which preserve visible content, layout, and multimodal cues while providing more consistent content-level observations for the same fetched source. This allows acquired information to be identified and compared across rollout trajectories, making visual webpage representation a basis for downstream credit assignment rather than merely an input modality choice.
% Building upon this representation, we propose Information-Aware Credit Assignment (ICA), a post-hoc credit attribution approach tailored to long-horizon retrieval. Instead of scoring actions online during rollout, ICA operates on completed trajectories to estimate the counterfactual contribution of each retrieved atomic evidence unit to the terminal outcome via posterior success association, and subsequently propagates the resulting dense learning signal to the upstream search turns responsible for acquiring high-utility information. We define an atomic evidence unit as the minimal identifiable unit of external information acquired through tool interactions; concretely, under web search it corresponds to an individual result item, whereas under fetch it corresponds to a retrieved webpage snapshot.
% To make evidence attribution feasible, we represent fetched webpages as rendered snapshots. A snapshot preserves the visible content of a page together with its layout, tables, images, and other multimodal cues, allowing the agent to observe the webpage in a form closer to how information is presented on the open web. More importantly, snapshots provide a consistent content-level observation for the same fetched source, which makes acquired information easier to identify and compare across rollout trajectories. In this view, visual webpage representation is not merely an input modality choice, but a way to construct stable information units that can support downstream credit assignment.

Finally, as illustrated in Figure \ref{fig:pipeline}, we incorporate snapshots and ICA into a GSPO-based \cite{zheng2025groupsequencepolicyoptimization} optimization pipeline and evaluate the resulting agents on a diverse suite of information-seeking benchmarks. Empirically, our approach yields consistent gains over text-based and outcome-only baselines across different model configurations, underscoring the value of grounding reinforcement learning in visually structured external observations and performing credit assignment at the granularity of acquired evidence.

In summary, this work makes the following contributions:
\begin{itemize}
    % \item We introduce rendered webpage snapshots as a stable, information-preserving modality for external observations, facilitating stable identification and cross-trajectory comparison of acquired evidence without lossy, model-dependent preprocessing.

    % \item We propose Information-Aware Credit Assignment (ICA), a post-hoc, information-level credit assignment mechanism that estimates the utility of acquired information via posterior success association and propagates it back to the decisions that introduced it.

    % \item We integrate snapshots and ICA into a GSPO-based optimization framework and demonstrate consistent improvements on a diverse set of long-horizon information-seeking benchmarks compared to prior work.
    \item We formulate long-horizon web agent learning as an evidence attribution problem, where retrieval actions are optimized through the information they introduce rather than only through terminal answer outcomes.
    \item We introduce rendered webpage snapshots as stable content-level observations, enabling more consistent identification and comparison of fetched information across trajectories.
    \item We propose Information-Aware Credit Assignment, a post hoc credit assignment method that estimates association-based utility scores for tool-acquired information and propagates dense rewards to the search and fetch turns that introduced it.
    \item We integrate ICA with GSPO and show consistent improvements on BrowseComp, GAIA, Xbench-DS, and Seal-0, with our 8B model outperforming most reported open-source agents under 15B parameters.
\end{itemize}

\section{Related Work}
\subsection{Information Seeking Agents}

Information seeking agents have advanced rapidly with large language models and external tool use. Proprietary systems such as DeepResearch \citep{openai2025deepresearch_system_card} and Grok 3 \citep{xai2025grok3} have shown strong capabilities in open web search and complex question answering, although their training data, tool interfaces, and optimization procedures remain largely undisclosed. In parallel, open-source agents such as WebDancer \citep{wu2026webdancer}, WebThinker \citep{NEURIPS2025_ae03bdef}, R1 Searcher \citep{song2025r1searcherincentivizingsearchcapability}, and related systems adopt ReAct style interaction patterns and provide reproducible platforms for studying multi-turn search, evidence acquisition, and answer generation.

A central challenge in these systems is how external web information is represented and maintained during search. Some agents interact with the web through browsing operations such as search, click, and open, where clicked or opened pages are returned as textual webpage content \citep{lu2025deepdiveadvancingdeepsearch}. Such text observations often contain boilerplate, navigation elements, and formatting noise, which consume large context budgets. Search and summary pipelines compress webpages before reasoning \citep{tongyideepresearchteam2026tongyideepresearchtechnicalreport}, but summarization may remove details that later become important and introduce model-dependent variation across trajectories. Retrieval-based pipelines such as Search R1 \citep{jin2025searchr1} improve efficiency by chunking webpages and selecting relevant segments, yet their performance depends on chunk boundaries and reranking quality. These approaches make web information easier to consume, but they usually treat retrieved content as transient text inputs rather than identifiable information units that can be compared across trajectories. In contrast, our work represents fetched webpages as rendered snapshots and uses tool-acquired information units as the basis for assigning credit to retrieval decisions.

\subsection{Reinforcement Learning in IS Agents}

Reinforcement learning has reshaped large language models through RLHF \cite{ziegler2019finetuning,stiennon2022learningsummarizehumanfeedback,ouyang2022training,rafailov2023direct}. More recently, group-based RL methods such as GSPO \cite{zheng2025groupsequencepolicyoptimization}, Dr. GRPO \cite{liu2025understanding}, and DAPO \cite{yu2026dapo} have achieved strong performance without learning an explicit value function. However, applying RL to information seeking agents remains challenging because long trajectories make outcome-only supervision sparse and noisy.

Recent work has explored denser credit assignment for multi-turn agents. IGPO \cite{wang2026information} rewards each turn by its marginal gain in correct answer probability, MT PPO \cite{wei2025reinforcingmultiturnreasoningllm} uses an LLM judge to score individual steps, and GiGPO \cite{feng2026group} contrasts actions through anchor state grouping. These methods reduce reward sparsity, but they mainly assign credit through turns, states, or answer probability changes. In web-based information seeking, the value of a tool action depends on the external information it introduces. ICA instead assigns credit through acquired information units and propagates the resulting rewards back to the search and fetch actions that introduced useful sources or content.

\subsection{Information Seeking Benchmark}
The landscape of information-seeking tasks has evolved from basic single-hop queries to complex, multi-hop reasoning, and eventually to highly nonlinear, deep-search challenges. Early single-hop benchmarks, such as NQ~\cite{kwiatkowski-etal-2019-natural}, TriviaQA \cite{joshi-etal-2017-triviaqa}, and SimpleQA-Verified \cite{haas2025simpleqa}, primarily consist of questions resolvable through structured queries, simple search operations, or even the model's internal parametric knowledge. Subsequent multi-hop datasets, including Bamboogle \cite{press2023measuring}, Xbench-DS \cite{xbench2025deepsearch}, and GAIA \cite{mialon2024gaia}, have significantly elevated the difficulty by introducing multi-step retrieval requirements and domain-specific annotations. These benchmarks rigorously test an agent's capability in deep search and multi-turn tool utilization. At the frontier of this complexity are benchmarks like BrowseComp \cite{wei2025browsecomp} and Seal-0 \cite{pham2026sealqaraisingbarreasoning}, which represent the most formidable challenges for information-seeking agents. These tasks are characterized by highly nonlinear coupled sub-problems and deliberate information masking, introducing substantial uncertainty and search depth that push the boundaries of current agentic reasoning.

\section{Preliminaries}

% \label{sec:problem-formulation}

\newcommand{\Search}{\textsc{Search}}
\newcommand{\Fetch}{\textsc{Fetch}}
\newcommand{\Answer}{\textsc{Answer}}  
\newcommand{\History}{\mathcal{H}}
\newcommand{\QuerySet}{\mathcal{Q}}
\newcommand{\ActionSet}{\mathcal{A}}

We consider an IS agent that answers a user query 
$q \in \mathcal{Q}$ through sequential interactions with external web resources. 
The agent incrementally acquires external information, updates its interaction history, 
and produces a final response conditioned on the accumulated evidence.

\paragraph{State.}
At turn $t$, the agent state is defined as the cumulative interaction history
\begin{equation}
  s_t = \mathcal{H}_{t-1}
  =
  \left(q, \xi_1, \ldots, \xi_{t-1}\right),
\end{equation}
where $q$ is the original user query and each interaction record is defined as
\begin{equation}
  \xi_t = (r_t, a_t, o_t).
\end{equation}
Here, $r_t$ denotes the agent's intermediate reasoning content, $a_t$ denotes the executable action parsed from the agent's output, and $o_t$ denotes the corresponding external observation at turn $t$.

\paragraph{Action space and interaction protocol.}
At turn $t$, the agent chooses an action
\begin{equation}
  a_t \in \mathcal{A}=\{\Search,\Fetch,\Answer\}.
\end{equation}
A \Search{} action issues one or more queries to the web environment, which returns a ranked list of search results
\begin{equation}
  o_t^{\mathrm{search}}=\{u_{t,k}\}_{k=1}^{n},
\end{equation}
where each result contains a URL, a title, and a short snippet. A \Fetch{} action selects a subset of previously observed URLs, denoted by $L_t$, and retrieves their corresponding contents:
\begin{equation}
  o_t^{\mathrm{fetch}}=\{\mathcal{C}(\ell):\ell\in L_t\}.
\end{equation}
Here, $\mathcal{C}(\ell)$ denotes the information returned from URL $\ell$, instantiated as either parsed webpage text or a rendered page snapshot.

The interaction history is updated sequentially as
\begin{equation}
  \mathcal{H}_t = \mathcal{H}_{t-1} \oplus (r_t, a_t,o_t),
\end{equation}
where $\oplus$ denotes sequence concatenation. The process terminates when the agent selects \Answer, after which the final response is generated as
\begin{equation}
  a_T=\Answer, 
  \qquad 
  \hat{y}\sim \pi_\theta(\cdot\mid \mathcal{H}_{T-1}).
\end{equation}

\section{Information-Aware Credit Assignment for Information-Seeking Agents}

% \subsection{Information-Aware Credit Assignment (ICA)}
% \label{sec:eica}

To address the challenges above, we propose \emph{Information-Aware Credit Assignment (ICA)}, 
a turn-level credit assignment mechanism for long-horizon information-seeking agents. 
ICA identifies which action turns are useful for solving the task by examining whether they acquire informative external evidence, 
and assigns learning signals to those turns instead of uniformly propagating the final outcome reward across the trajectory.

\subsection{Turn-Level Credit Assignment}
\label{sec:turn-reward}

Given a prompt $x$, we sample a group of trajectories 
$G=\{\tau_1,\ldots,\tau_N\}$ from $\pi_{\theta_{\mathrm{old}}}$, 
where each trajectory receives a binary outcome reward $R_i\in\{0,1\}$. 
To address the long-horizon credit assignment challenge, ICA estimates the utility of URLs 
$U_t^{i}$ acquired at each action turn $a_t^{i}$. 
The URLs are then ranked by their utility scores, and the highest-ranked ones are selected as useful evidence units.
Tool turns containing any selected evidence unit receive positive credit, while the final answer turn is supervised by the trajectory-level outcome reward. 
For failed trajectories, ICA further favors non-terminal exploratory turns over premature answer attempts, encouraging the agent to continue evidence acquisition when the available information is insufficient.
\paragraph{Atomic-Evidence Association-Based Utility}
\label{sec:atomic-evidence-counterfactual}

We define an atomic evidence unit $e$ as the minimal identifiable unit of external information that can be acquired and reused by the agent. Let $\mathcal{E}$ denote the union of all atomic evidence units acquired across trajectories in the rollout group. In our implementation, an atomic evidence unit corresponds to a stable URL-level evidence item, specifically an individual result entry returned by \Search{} or a complete webpage snapshot acquired by \Fetch{}.

For each completed trajectory $\tau_i$, let $\mathcal{E}_{i} \subseteq \mathcal{E}$ denote the set of atomic evidence units in the $i$-th trajectory. Then, the acquisition
indicator is defined as
\begin{equation}
  I_e^{i} \;=\; \mathbb{I}\!\left[e \in \mathcal{E}_{i}\right] \in \{0,1\}.
\end{equation}
Each trajectory is associated with a binary outcome $R_i \in \{0,1\}$
indicating task success.

We estimate the empirical success probability conditioned on acquiring $e$ as
\begin{equation}
  P_e^+ = \frac{\sum_{i=1}^{N} I_e^{i} R_i}{\sum_{i=1}^{N} I_e^{i}} ,
\end{equation}
and similarly, the success probability conditioned on not acquiring $e$ as
\begin{equation}
  P_e^- = \frac{\sum_{i=1}^{N} \left(1-I_e^{i}\right) R_i}{\sum_{i=1}^{N} \left(1-I_e^{i}\right)} .
\end{equation}
When the denominator of either estimator is zero, we fall back to the
batch-level success rate
$P_e^+ = P_e^- =\frac{1}{N}\sum_{i=1}^{N} R_i$
to ensure numerical stability.

The association-based utility of $e$ is then defined as
the difference between these two conditional success probabilities:
\begin{equation}\label{eq:delta_e}
  \Delta_e = P_e^+ - P_e^- .
\end{equation}

We compute $\Delta_e$ only for groups with mixed outcomes, i.e., groups in which at least one trajectory succeeds and at least one trajectory fails. If all trajectories in a group have the same outcome, the group provides no contrastive signal for tool-level credit assignment, and all tool-turn advantages are set to zero. Importantly, $\Delta_e$ is used only to rank atomic evidence units, rather than being used directly as a continuous reward. In subsequent ablation studies, we empirically demonstrate a strong correlation between $\Delta_e$ and answer correctness, validating its efficacy as a proxy for information utility.

For each tool type $c \in \{\Search,\Fetch\}$, let $\mathcal{E}^{c}$ denote the set of atomic evidence units acquired by tool type $c$ within the group. We rank atomic evidence units according to their association-based utility score $\Delta_e$, and select the top positive units:
\begin{equation}
\mathcal{T}^{c} = \bigl\{ e \in \mathcal{E}^{c} \;\big|\; \mathrm{rank}(e) \le k_c \,\land\, \Delta_e > 0 \bigr\} ,
\end{equation}
where
\begin{equation}
k_c = \max\Bigl(1, \,\bigl\lfloor \rho |\mathcal{E}^{c}| \bigr\rfloor\Bigr) .
\end{equation}
Here, $\rho \in (0, 1)$ is a hyperparameter governing the top fraction of retrieved URLs to be designated as high-utility evidence, and $\mathrm{rank}(e)$ denotes the descending rank order of $e$ based on its association-based utility score $\Delta_e$.

% \paragraph{Action-Turn Reward}
% \label{sec:binary-tool-turn-reward}

% Let $\mathcal{E}_{j}^{(n)} \subseteq \mathcal{E}$ denote the set of atomic
% evidence units acquired at the $j$-th action turn of trajectory $\tau^{(n)}$, and
% let $c_j$ denote the corresponding tool type. Instead of using the raw
% counterfactual score as a continuous reward, we assign a binary reward according
% to whether the turn acquires at least one selected atomic evidence unit:
% \begin{equation}
% r_j^{(n)}
% =
% \begin{cases}
% 1, & \text{if } \mathcal{E}_{j}^{(n)} \cap \mathcal{T}^{c_j} \neq \emptyset,\\
% 0, & \text{otherwise}.
% \end{cases}
% \end{equation}

% \paragraph{Answer Turn Reward}
% For the final answer turn, we use an asymmetric shaped reward. Let $| \tau^{(n)} |$
% denote the number of generated tokens in trajectory $\tau^{(n)}$, and let
% $L_{\max}$ denote the maximum context window. The shaped answer reward is
% \begin{equation}
% \small
% \tilde{R}^{(n)}
% =
% \begin{cases}
% 1.0, & R^{(n)} = 1,\\
% -1.0 + \sqrt{\frac{|\tau^{(n)}|}{L_{\max}}},
% & R^{(n)} = 0.
% \end{cases}
% \end{equation}
\paragraph{Action-Turn Reward}
To decouple search quality from the final outcome, we assign distinct rewards to intermediate tool turns and the terminal answer turn. For any intermediate action turn $t$ invoking a tool of type $c_t^{i} \in \{\Search, \Fetch\}$, we evaluate its acquired URL set $\mathcal{E}_{t}^{i}$. Instead of relying on association-based utility scores that easily collapse under small group sizes, we assign a binary reward $r_t^{i}$:
\begin{equation}
  r_t^{i} = \mathbb{I}\bigl[ \mathcal{E}_{t}^{i} \cap \mathcal{T}^{c_t^{i}} \neq \emptyset \bigr] .
\end{equation}
This means the tool turn is explicitly rewarded with $1$ if it successfully captures at least one high-utility evidence unit from $\mathcal{T}^{c_t^{(i)}}$, and $0$ otherwise. Conversely, the terminal answer turn is evaluated via an asymmetric shaped reward $\tilde{R}^{(n)}$:
\begin{equation}
  \tilde{R}_i = 
  \begin{cases}
    1.0, & \text{if } R_i = 1, \\
    -1.0 + \sqrt{\frac{|\tau_i|}{L_{\max}}}, & \text{if } R_i = 0.
  \end{cases}
\end{equation}
Here, $R_i \in \{0, 1\}$ indicates the correctness of the final answer. This budget-aware shaping mechanism scales down the penalty based on the consumed token count $|\tau_i|$ relative to the maximum context window $L_{\max}$, thereby protecting thorough search exploration from being heavily penalized upon failure.

\paragraph{Turn-Level Advantage Assignment}
\label{sec:turn-advantage-assignment}

To explicitly decouple search-quality supervision from final-answer outcomes, we assign advantages by centering the rewards strictly within their respective scopes. Each token generated at turn $t$ in trajectory $\tau_i$ inherits the turn-level advantage $A_t^{i}$:
\begin{equation}
A_t^{i} = 
\begin{cases}
\frac{r_t^{i} - \bar{r}}{\sigma_r}, & a_t^i \in \{\Search, \Fetch\}, \\
\frac{\tilde{R}_i - \bar{R}}{\sigma_R}, & a_t^i \in \{\Answer\}.
\end{cases}
\end{equation}
Here, the baselines $\bar{r}$ and $\bar{R}$ are the group-level mean rewards for the specific tool type $c_t^i$ and the final answer, respectively, while $\sigma_r$ and $\sigma_R$ denote their corresponding standard deviations. Specifically, for the intermediate tool calls, we compute $\bar{r} = \frac{1}{|\mathcal{S}^c|} \sum_{(i,t) \in \mathcal{S}^c} r_i^{t}$, where $\mathcal{S}^{c}$ denotes the set of all tool turns of type $c$ within the group. 
\paragraph{ICA-GSPO Optimization}
\label{sec:ica-gspo}

We optimize the policy using a turn-level GSPO objective. Building upon the trajectories and advantages established in Sec.~\ref{sec:turn-advantage-assignment}, action turns receive information-acquisition advantages, and the final answer turn receives the outcome-level advantage. 

Unlike standard GSPO which computes a single importance ratio over the entire trajectory, ICA-GSPO computes this ratio independently within each turn, inspired by recent advances in turn-level importance sampling \citep{li2026stabilizingoffpolicytraininglonghorizon}. This design aligns with the natural multi-turn structure of the agent and prevents large policy shifts in a single token from contaminating the trajectory-level estimate. For the $t$-th turn of trajectory $\tau_i$, all tokens share a common geometric-mean importance ratio $s_t^{i}(\theta)$:
\begin{equation}
s_t^{i}(\theta) = \exp \left( \frac{1}{|\tau_t^{i}|} \sum_{l \in \tau_t^{i}} \log \frac{\pi_{\theta}(y_l \mid \mathcal{H}_{t-1}^i, y_{<l})}{\pi_{\theta_{\mathrm{old}}}(y_l \mid \mathcal{H}_{t-1}^i, y_{<l})} \right) .
\end{equation}

Using this turn-level ratio and the normalized advantage $A_t^i$, we define the turn-level clipped GSPO surrogate objective as
\begin{equation}
\begin{aligned}
\mathcal{J}_t^{i}(\theta)
= \min \Big\{
& s_t^{i}(\theta)A_t^{i}, \\
& \mathrm{clip}\big(s_t^{i}(\theta),
1-\epsilon,1+\epsilon\big)
A_{t}^{i}
\Big\}.
\end{aligned}
\end{equation}
To prevent the critical learning signal of the single answer turn from
being diluted by a potentially large number of preceding tool turns,
we aggregate the surrogate objectives of the non-answer turns and the
final answer turn using a fixed weighting factor $w$:
\begin{equation}
\begin{aligned}
\mathcal{J}_i(\theta)
=
&\, w\,\mathcal{J}_{T_i}^{i}(\theta) \\
&+
(1-w)\frac{1}{T_i-1}
\sum_{k=1}^{T_i-1}
\mathcal{J}_{k}^i(\theta).
\end{aligned}
\end{equation}

The overall objective is then obtained by taking the expectation of the
trajectory-level surrogate objective over trajectories sampled from the old policy:
\begin{equation}
\mathcal{J}_{\mathrm{ICA\text{-}GSPO}}(\theta)
=
\mathbb{E}_{\tau \sim \pi_{\theta_{\mathrm{old}}}}
\left[
\mathcal{J}(\theta;\tau)
\right].
\end{equation}

\subsection{External Observations}
% \subsubsection{Webpage Snapshots as Observations}
% Information-aware credit assignment relies on stable and information-complete external observations, as credit is assigned post-hoc after trajectory completion.
Existing approaches commonly obtain web content in text-only form, either by summarizing webpages or retrieving a subset of passages via RAG-style pipelines \cite{jin2025searchr1}. Such representations introduce irreversible information loss at acquisition time, break cross-trajectory consistency for the same information source, and inject additional model bias through learned summarization or retrieval components. These properties are fundamentally misaligned with ICA, which estimates the utility of external information units based on their aggregate contribution to successful outcomes across trajectories.

\begin{figure}[h]
  \centering

  \begin{subfigure}{0.95\linewidth}
    \centering
    \includegraphics[width=\linewidth]{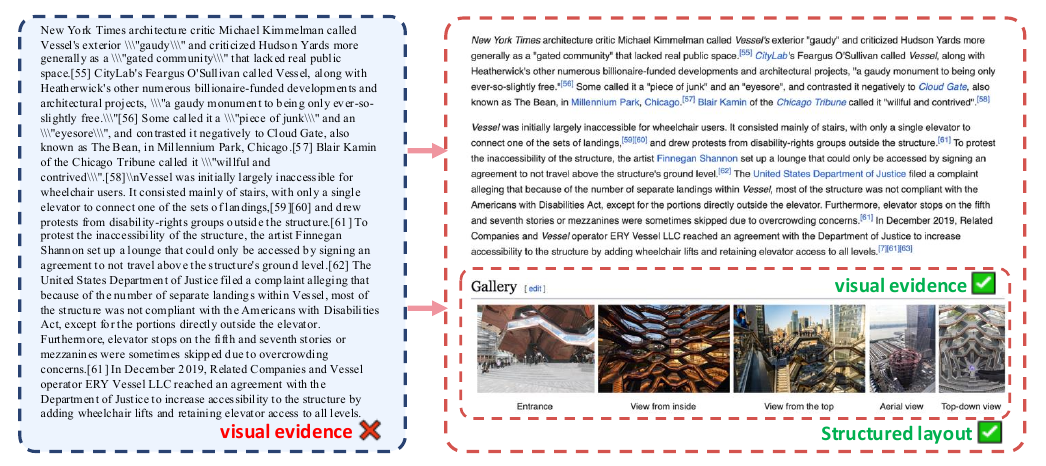}
    \caption{Full Text \& Snapshot-Based Representation}
    \label{fig:top}
  \end{subfigure}

  \vspace{0.6em} % 控制上下间距（可删/可调）

  \begin{subfigure}{0.95\linewidth}
    \centering
    \includegraphics[width=\linewidth]{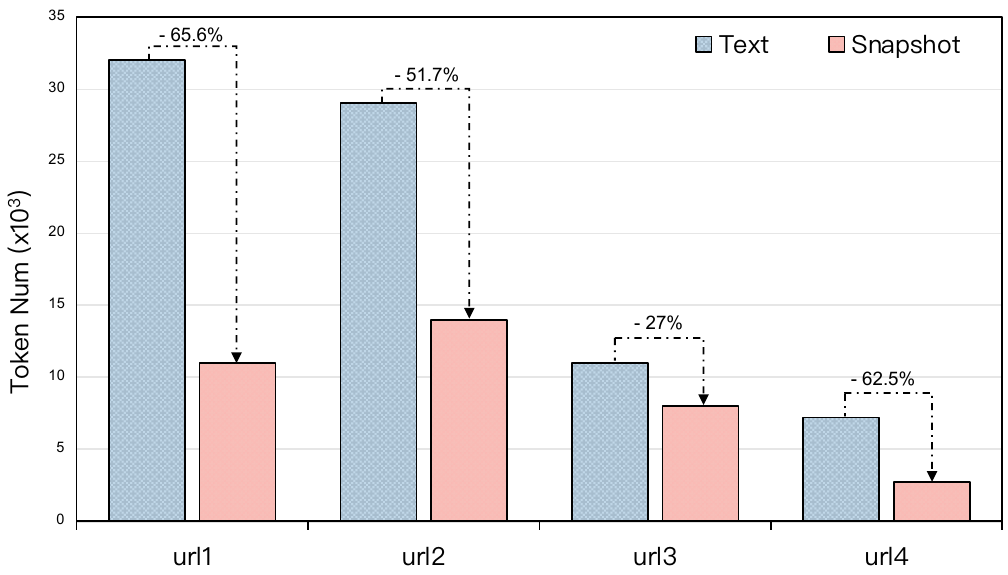}
    \caption{Token Budget Comparison}
    \label{fig:bottom}
  \end{subfigure}

  \caption{ Text vs Visual snapshots. \textbf{(a)} Plain text extracted from webpages is often cluttered and lacks visual structure, while snapshots preserve layout and non-textual cues. \textbf{(b)} Token count comparison across four URLs, showing that snapshots reduce token usage by 27.0\% -- 65.6\% compared to parsed text.}
  \label{fig:example2}
  \vspace{-1em}
\end{figure}

To address this issue, we model external observations as webpage snapshots rather than summarized or retrieved text. A snapshot preserves the full visible content of a webpage, including textual elements, layout structure, tables, and images, and is treated as a persistent external information unit in the agent's memory. Snapshots are returned directly by tool invocations and consumed by the agent without intermediate filtering or compression, allowing the agent to determine which information becomes relevant during downstream reasoning, as illustrated in Figure \ref{fig:example2}. This representation provides a stable and consistent basis for identifying recurring information units across trajectories.

% Beyond preserving information fidelity, webpage snapshots offer several practical advantages for long-horizon information-seeking.
% First, they prevent premature information pruning, ensuring that potentially useful evidence remains available for post-hoc credit assignment even if it is only utilized at later reasoning stages.
% Second, despite retaining richer structure, snapshots are often more token-efficient than raw webpage text, as visual layout naturally suppresses boilerplate content such as navigation bars and repeated footers.
% Figure[add the image] illustrates that snapshot-based observations consistently require fewer tokens than fetching full webpage text across diverse websites.
% Third, snapshots provide consistent representations for the same URL across different trajectories, supporting reliable cross-trajectory utility estimation.
% Finally, as snapshots persist across turns, they naturally support multi-hop reasoning and cumulative evidence aggregation, closely reflecting how humans interact with web information.
% Together, these properties make webpage snapshots a well-suited observation modality for information-aware credit assignment in volatile web environments.

% \subsubsection{Case Study}

\section{Experiments}

\begin{table*}[ht]
\centering
\scriptsize
\setlength{\tabcolsep}{5pt} % 稍微减小列间距
\caption{Comparison of different models on information-seeking benchmarks. We report the Pass@1 accuracy on BrowseComp, GAIA, Xbench-DS, and Seal-0, while ``snap.'' refers to visual snapshot-based fetching. Results marked with $^{*}$ are taken from existing studies. Best open-source results are highlighted in \textbf{bold}.}
\label{tab:model_comparison_final_tools}
\begin{tabularx}{\linewidth}{l >{\centering\arraybackslash}p{3cm} *{4}{>{\centering\arraybackslash}X}}
\toprule
\textbf{Model / Framework} & \textbf{Tools} & \textbf{BrowseComp} & \textbf{GAIA} & \textbf{Xbench-DS} & \textbf{Seal-0} \\
\midrule
% --- Proprietary Agents ---
\rowcolor{bg_green}
\multicolumn{6}{l}{\textit{\textbf{Proprietary Agents}}} \\
Claude-4-Sonnet$^{*}$ & -- & 12.2 & 68.3 & 64.6 & -- \\
OpenAI-o3$^{*}$ & -- & 49.7 & 70.5 & 66.7 & 18.9 \\
OpenAI DeepResearch$^{*}$ & -- & 51.5 & 67.4 & -- & -- \\
\midrule
% --- Open-Source Agents (<15B) ---
\rowcolor{bg_pink}
\multicolumn{6}{l}{\textit{\textbf{Open-Source Agents ($<$15B)}}} \\
WebExplorer-8B$^{*}$ & Search \& Fetch (text) & \uline{15.7} & \uline{50.0} & \uline{53.7} & -- \\
WebSailor-7B$^{*}$ & Search \& Fetch (text) & 6.7 & -- & 34.3 & -- \\
DeepDive-9B$^{*}$ & Search \& Fetch (text) & 6.3 & -- & 38.0 & \uline{12.2}\\
MiroThinker-14B-DPO-v0.1$^{*}$ & Search \& Fetch (text) & 9.0 & -- & 30.0 & -- \\
Qwen3-14B$^{*}$ & Search \& Fetch (text) & 1.0 & -- & 20.0 & -- \\
% InfoAgent-SFT$^{*}$ & Search \& Fetch (text) & 4.7 & -- & 28.0 & -- \\
InfoAgent$^{*}$ & Search \& Fetch (text) & 15.3 & -- & 40.4 & -- \\

% --- Ours (<15B) ---
\rowcolor{bg_blue} Qwen3-VL-8B-ICA (Ours) & Search \& Fetch (snap.) & \textbf{25.0} & \textbf{69.9} & \textbf{71.0} & \textbf{25.2} \\
\midrule
% --- Open-Source Agents (>15B) ---
\rowcolor{bg_pink}
\multicolumn{6}{l}{\textit{\textbf{Open-Source Agents ($>$15B)}}} \\
ASearcher-Web-32B$^{*}$ & Search \& Fetch (text) & 5.2 & 52.8 & 42.1 & -- \\
MiroThinker-32B-DPO-v0.2$^{*}$ & Search \& Fetch (text) & 13.0 & 64.1 & -- & -- \\
Kimi-K2-Instruct-1T$^{*}$ & Search \& Fetch (text) & 14.1 & 57.7 & 50.0 & -- \\
WebDancer-QwQ-32B$^{*}$ & Search \& Fetch (text) & 3.8 & 51.5 & 38.3 & -- \\
WebSailor-32B$^{*}$ & Search \& Fetch (text) & 10.5 & 53.2 & 53.3 & 21.3 \\
DeepDive-32B$^{*}$ & Search \& Fetch (text) & 15.3 & -- & 51.8 & 25.5 \\
C-GSPO$^{*}$ & Search \& Fetch (text) & \uline{24.8} & 56.3 & 57.7 & -- \\
WebShaper-QwQ-32B$^{*}$ & Search \& Fetch (text) & -- & 53.3 & 35.0 & -- \\
WebLeaper$^{*}$ & Search \& Fetch (text) & 22.7 & \uline{69.9} & \uline{62.3} & \textbf{35.1}  \\
% --- Ours (>15B) ---
\rowcolor{bg_blue} Qwen3-VL-30B-A3B-ICA (Ours) & Search \& Fetch (snap.) & \textbf{26.1} & \textbf{72.8} & \textbf{76.0} &  \uline{27.0}\\
\bottomrule
\end{tabularx}
\end{table*}

\begin{table}[ht]
\centering
\caption{Ablation study of model components. \textbf{BC-100} denotes a subset of 100 queries randomly sampled from BrowseComp, and \textbf{XDS} denotes Xbench-DS. All results are reported as Pass@1 accuracy.}
\label{tab:Ablation_information_seeking}
\renewcommand{\arraystretch}{0.95} % 稍微拉大一点行距，视觉更舒适
\setlength{\tabcolsep}{4pt} % 调整列间距

% 使用 resizebox 将表格完美缩放至 \linewidth，需确保导言区有 \usepackage{graphicx}
\resizebox{\linewidth}{!}{
\begin{tabular}{lcccc}
\toprule
\textbf{Method} & \textbf{BC-100} & \textbf{GAIA} & \textbf{XDS} & \textbf{Seal-0}\\
\midrule

\multicolumn{5}{l}{\textbf{Baseline: Qwen3-VL-8B-Thinking}}\\
\midrule
Baseline - Text  & 1.0 & 29.1 & 39.0 & 7.2\\
SFT - Text   & 11.0 \textcolor{mygreen}{(+10.0)} & 54.3 \textcolor{mygreen}{(+25.2)} & 55.0 \textcolor{mygreen}{(+16.0)} & 20.7 \textcolor{mygreen}{(+13.5)}\\
SFT - Snap. & 18.0 \textcolor{mygreen}{(+17.0)} & 56.4 \textcolor{mygreen}{(+27.3)} & 66.0 \textcolor{mygreen}{(+27.0)} & 23.4 \textcolor{mygreen}{(+16.2)}\\
 GSPO - Snap. & 22.0 \textcolor{mygreen}{(+21.0)} & 60.2 \textcolor{mygreen}{(+31.1)} & 68.0 \textcolor{mygreen}{(+29.0)} & 24.3 \textcolor{mygreen}{(+17.1)}\\
ICA - Snap. & 25.0 \textcolor{mygreen}{(+24.0)}  & 69.9 \textcolor{mygreen}{(+40.8)}& 71.0 \textcolor{mygreen}{(+32.0)} & 25.2 \textcolor{mygreen}{(+18.0)}\\
\bottomrule
% \multicolumn{6}{l}{\textbf{Baseline: Qwen3-VL-30B-A3B-Thinking}}\\
% \midrule
% \textit{Baseline}  & Base - RAG   & 3.0  & 31.1 & 38.0 & 9.9\\
% \multirow{2}{*}{\textit{SFT}}
%                   & SFT - RAG    & 10.0 \textcolor{mygreen}{(+7.0)} & 57.3 \textcolor{mygreen}{(+26.2)} & 61.0  \textcolor{mygreen}{(+23.0)}& 22.5 \textcolor{mygreen}{(+12.6)}\\
%                   & SFT - Snap.  & 11.0 \textcolor{mygreen}{(+8.0)} & 60.2 \textcolor{mygreen}{(+29.1)} & 64.0 \textcolor{mygreen}{(+26.0)} & 23.4 \textcolor{mygreen}{(+13.5)}\\
% \multirow{2}{*}{\textit{RL}}
%                   & GSPO - Snap. & 13.0 \textcolor{mygreen}{(+10.0)} & 57.3 \textcolor{mygreen}{(+26.2)}& 66.0 \textcolor{mygreen}{(+28.0)} & 24.3 \textcolor{mygreen}{(+14.4)}\\
%                   & ICA - Snap.  & 17.0 \textcolor{mygreen}{(+14.0)} & 65.0 \textcolor{mygreen}{(+33.9)}& 75.0 \textcolor{mygreen}{(+37.0)}& 27.0 \textcolor{mygreen}{(+17.1)}\\

% \bottomrule
\end{tabular}
}
\end{table}

\subsection{Main Results}

This section presents the main empirical results of our approach.
We primarily evaluate Information-Aware Credit Assignment on a suite of
long-horizon \textbf{information-seeking benchmarks}, comparing against both
proprietary and open-source agentic systems to assess performance in dynamic
web environments that require iterative search, evidence acquisition, and
multi-step reasoning.

\subsubsection{Performance on Information Seeking Benchmarks}

Table~\ref{tab:model_comparison_final_tools} compares our approach with proprietary and open-source agentic systems on long-horizon information-seeking benchmarks. Many text-based baselines in Table 1 use summarized, reranked, or retrieval-filtered fetch results rather than raw webpage dumps. Thus, our comparison is not merely against naive text parsing, but against commonly used compressed text-fetching pipelines for web agents. Overall, ICA demonstrates strong and consistent performance across all evaluated tasks, validating the effectiveness of Information-Aware Credit Assignment in complex, multi-step web environments.

% Among open-source methods, \textbf{ICA-32B} achieves the best average
% success rate (\textbf{54.41}), outperforming strong baselines such as WebLeaper
% (48.3) and C-GSPO (43.0), and attaining the highest scores on GAIA (83.5) and
% Xbench-DS (73.21).
% At the smaller scale, \textbf{ICA-8B} remains competitive, achieving an
% average success rate of 44.41 and matching or surpassing several baselines of
% similar or larger sizes.

Compared with proprietary systems, ICA-30B-A3B remains competitive on
challenging benchmarks such as BrowseComp and Seal-0, despite using a
significantly smaller model size and a fully open training pipeline. 
% In addition, we report a diagnostic evaluation on \textbf{Bamboogle}, a
% multi-hop question answering benchmark with a shorter reasoning horizon.
% On Bamboogle, \textbf{ICA-8B} achieves a success rate of \textbf{80.8\%},
% while \textbf{ICA-30B-A3B} further improves performance to \textbf{91.2\%},
% indicating that our method provides effective learning signals even in
% relatively short-horizon search settings.

\subsubsection{Ablation Study}
\label{sec:ablation}

Table~\ref{tab:Ablation_information_seeking} isolates the effects of (i) the fetch representation used in SFT
(text parsing vs. Snapshot visual grounding) and (ii) the RL optimization scheme (comparing GSPO with ICA).
\paragraph{Comparison of Snapshot and Text in SFT.}
Across both backbones, adopting snapshot-based fetches in SFT yields consistent improvements over textual fetches, with gains concentrated on benchmarks that are more sensitive to layout and presentation
(e.g., GAIA and XDS). This indicates that snapshot observations better preserve layout-dependent semantics while reducing artifacts introduced by heuristic webpage parsing and text linearization, thereby enabling more reliable evidence
localization during multi-step retrieval.

\paragraph{Advantages of ICA in RL Optimization.}
With snapshot retrieval fixed as the fetch policy during RL, ICA consistently outperforms vanilla GSPO on all
benchmarks and both model scales. The improvements are particularly salient on long-horizon, high-noise tasks,
supporting the claim that information-aware post-hoc credit assignment provides a denser and more targeted
training signal by attributing outcome rewards to the retrieval turns that introduce high-utility evidence.
\begin{figure*}[t]
\centering
\includegraphics[width=\textwidth]{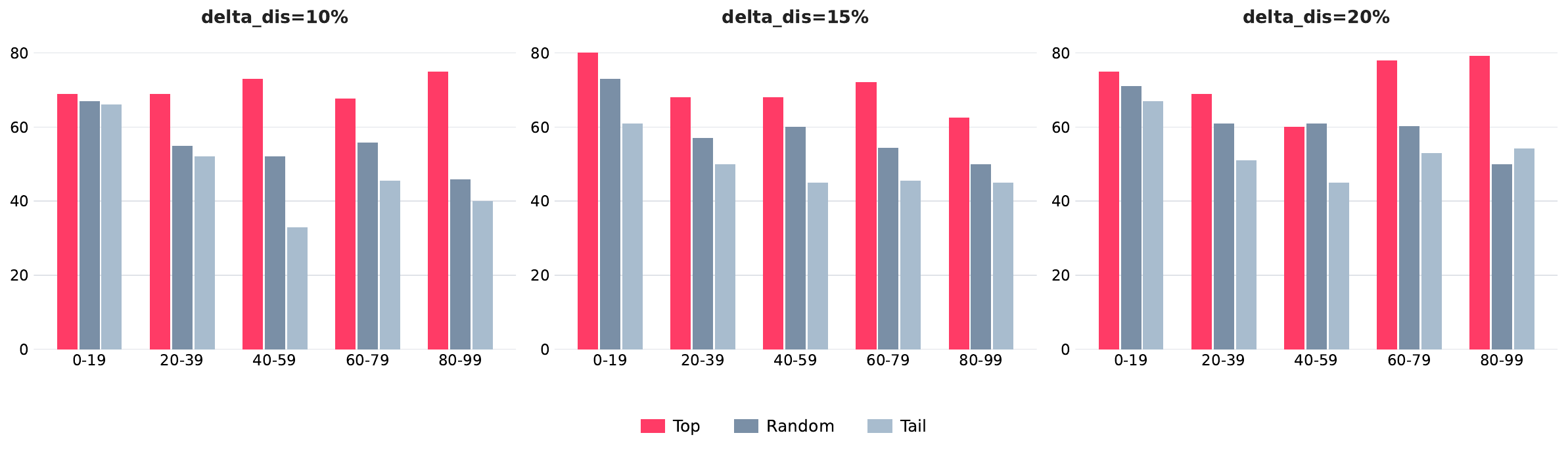}
\caption{Top-$p\%$ ICA evidence (red) consistently outperforms random and tail baselines across all trajectory lengths and thresholds ($p \in \{10, 15, 20\}$).}
\label{fig:topk_sufficiency}
\end{figure*}
\subsubsection{Analysis of ICA}

To further examine potential confounds and better understand the behavior of ICA, we conduct additional diagnostic experiments focusing on the quality of selected atomic evidence under controlled trajectory lengths.

\begin{table}[ht]
\centering
\caption{Evaluation under strict context constraints (128k context limit, 16k tool response truncation threshold). We report Pass@1 accuracy with the context overflow rate in parentheses. \textbf{XDS} denotes Xbench-DS.}
\label{tab:text_vs_snapshot}
\renewcommand{\arraystretch}{1.1} % 稍微增加一点行高，让包含括号的数据更舒展
\setlength{\tabcolsep}{8pt} % 因为列数减少了，可以适当增加列间距
\resizebox{\linewidth}{!}{
\begin{tabular}{l cccc}
\toprule
\textbf{Method} & \textbf{BC-100} & \textbf{GAIA} & \textbf{XDS} & \textbf{Seal-0} \\
\midrule
SFT - Text  & 11.0 (36.0) & 54.3 (15.5) & 55.0 (6.0) & 20.7 (20.0) \\
SFT - Snap. & \textbf{18.0} (\textbf{0.0}) & \textbf{56.4} (\textbf{0.0}) & \textbf{66.0} (\textbf{0.0}) & \textbf{23.4} (\textbf{0.0}) \\
\bottomrule
\end{tabular}
}
\end{table}
\noindent\textbf{Length-controlled sufficiency.}
To isolate the effect of trajectory length, we partition trajectories into five buckets based on the number of interaction steps: $0\text{--}19$, $20\text{--}39$, $40\text{--}59$, $60\text{--}79$, and $80\text{--}99$. Within each bucket, we evaluate the sufficiency of ICA-selected evidence using a dynamic \emph{top-$p\%$} protocol (where $p \in \{10, 15, 20\}$): we retain the top $p\%$ highest-scoring atomic evidence units according to ICA and prompt the model to answer using only these units. We compare this against two baselines: \\(i) randomly sampled evidence of the same proportion (\emph{random-$p\%$}) \\ (ii) the lowest-scoring evidence under ICA (\emph{tail-$p\%$}).

As shown in Figure~\ref{fig:topk_sufficiency}, the top-$p\%$ ICA evidence consistently achieves the highest answer accuracy across almost all trajectory lengths and filtering thresholds, with detailed values reported in Appendix Table~\ref{tab:length_controlled_sufficiency}. On average across these subsets, the ICA-selected evidence reaches an accuracy of 71.0\%, substantially outperforming both the random (58.2\%) and tail (50.2\%) baselines. These results indicate that ICA reliably assigns higher scores to atomic evidence that is more likely to be sufficient for answering the question. Importantly, this advantage holds across all trajectory buckets and selection ratios, suggesting that the effectiveness of ICA is not driven by a bias toward longer trajectories, but rather by its robust ability to identify answer-supporting evidence regardless of the search length.

\noindent\textbf{Token Efficiency and Context Overflow.} 
To investigate the context efficiency of different webpage representation modalities, we conduct a controlled evaluation under strict context length and token budget constraints. Specifically, we cap the maximum context window of the 8B model to 128k tokens. Furthermore, the token budget for each individual tool response is strictly bounded within a 16k window; any returning content exceeding this threshold is forcefully truncated. 

We report the Pass@1 accuracy alongside the context overflow rate (i.e., the percentage of trajectories that trigger context truncation) across four benchmarks in Table~\ref{tab:text_vs_snapshot}. Under identical context and tool response constraints, the Snapshot modality (SFT - Snapshot) consistently outperforms the raw text baseline (SFT - Text) in both task efficacy and token efficiency. Crucially, while the text modality suffers from severe context overflow, reaching 36.0\% on BrowseComp and 20.0\% on Seal-0, the snapshot modality has an overflow rate of only 0.0\% across all evaluated datasets. This drastically lower token footprint translates directly to substantial performance gains, boosting Pass@1 accuracy by 7.0\% on BrowseComp and 11.0\% on Xbench-DS. Notably, on GAIA, the snapshot modality improves Pass@1 accuracy from 54.3\% to 56.4\% while eliminating context overflow, whereas the text baseline still incurs a 15.5\% overflow rate. These results demonstrate that Snapshot web representations are inherently more compact than raw HTML text, effectively mitigating context bloat and enabling robust navigation within constrained budgets.

\section{Conclusion}
This work addresses two bottlenecks in long-horizon open-web information seeking: lossy text-derived webpage representations and sparse retrieval credit assignment. We represent fetched pages as rendered snapshots, yielding stable evidence units for cross-rollout comparison. Information-Aware Credit Assignment (ICA) then estimates association-based evidence utility from rollout success patterns and assigns dense rewards to retrieval turns that introduce high-utility information. Experiments on BrowseComp, GAIA, Xbench-DS, and Seal-0 show consistent gains, demonstrating the effectiveness of evidence-level credit assignment for training long-horizon web agents.
\newpage
\section*{Limitations}
\label{sec:limitations}

While our proposed method demonstrates strong performance on long-horizon information-seeking tasks, we acknowledge two primary limitations in our current design. 

First, regarding the webpage processing mechanism, our framework relies on uncompressed visual representations. When the agent fetches a webpage, the raw rendered snapshot is directly fed into the Vision Transformer (ViT) and converted into visual tokens. We currently do not apply any intermediate visual compression, spatial pooling, or salient region filtering prior to tokenization. Although this straightforward approach preserves the original visual layout and fine-grained multi-modal cues of the web environment, it inevitably introduces considerable token redundancy and computational overhead, particularly when processing visually dense or exceedingly long webpages.

Second, our system currently lacks a sophisticated context management strategy to handle unbounded exploration. As the agent continuously executes tools and accumulates evidence across multiple turns, the sequence length grows rapidly. Without dynamic context reduction techniques, the accumulated tokens can easily exceed the model's maximum context limit during continuous exploration and evidence accumulation, particularly when historical summarization, token eviction, or memory-based sliding windows are not applied. In such cases of context overflow, the trajectory must be forcefully truncated, which may lead to the loss of early-retrieved crucial evidence and constrain the agent's capability on tasks that require extensively deep search steps. Addressing these efficiency bottlenecks through visual token compression and adaptive memory management remains an important direction for future work.

% \section*{Impact Statement}
% This paper advances methods for training information-seeking agents by introducing Information-Aware Credit Assignment (ICA), which attributes outcome feedback to retrieved atomic evidence and propagates learning signals to the turns that introduced useful information. The primary intended impact is to improve the reliability and data-efficiency of agent learning in open-web settings, potentially benefiting applications such as question answering, research assistance, and domain-specific information discovery.
% \nocite{langley00}

\bibliography{custom}

%%%%%%%%%%%%%%%%%%%%%%%%%%%%%%%%%%%%%%%%%%%%%%%%%%%%%%%%%%%%%%%%%%%%%%%%%%%%%%%
%%%%%%%%%%%%%%%%%%%%%%%%%%%%%%%%%%%%%%%%%%%%%%%%%%%%%%%%%%%%%%%%%%%%%%%%%%%%%%%
% APPENDIX
%%%%%%%%%%%%%%%%%%%%%%%%%%%%%%%%%%%%%%%%%%%%%%%%%%%%%%%%%%%%%%%%%%%%%%%%%%%%%%%
%%%%%%%%%%%%%%%%%%%%%%%%%%%%%%%%%%%%%%%%%%%%%%%%%%%%%%%%%%%%%%%%%%%%%%%%%%%%%%%
\newpage
\appendix
\onecolumn
\section{Experimental Details}
We will provide a detailed description of the algorithmic workflow and the specific experimental setup in this section.
\subsection{Experiment Setup}
\paragraph{Environment Implementation}
For the environment implementation, we utilize the Serper API\footnote{\url{https://serper.dev}} for the \texttt{search} tool, while the \texttt{fetch} tool is instantiated in two modalities: a text-based baseline using Trafilatura\footnote{\url{https://github.com/adbar/trafilatura}} to retrieve parsed content, and our proposed snapshot-based setting using Playwright\footnote{\url{https://playwright.dev}} to capture full-page visual observations.

\paragraph{Benchmarks \& Metrics}
We evaluate our method on a suite of long-horizon information-seeking benchmarks
that require iterative search, evidence acquisition, and multi-step reasoning
in dynamic web environments.
Specifically, we adopt BrowseComp~\cite{wei2025browsecomp},
GAIA~\cite{mialon2024gaia}, Xbench-DS~\cite{xbench2025deepsearch},
and Seal-0~\cite{pham2026sealqaraisingbarreasoning} to assess agentic information-seeking capabilities
under varying task complexities and environmental dynamics.
For GAIA, we evaluate on the 103 examples from the text-only validation subset~\cite{li-etal-2025-search},
following standard evaluation protocols.
For Xbench-DS, we utilize the more mature 2505 version release for evaluation. Across all benchmarks, we report Pass@1 as the primary metric, obtained via
LLM-as-a-judge evaluation.

\paragraph{Baselines}
We compare against a diverse set of strong baselines spanning both proprietary
and open-source information-seeking agents.
These include proprietary systems such as Claude-4-Sonnet ~\cite{anthropic2025claude4}, OpenAI-o3 ~\cite{openai2025o3_o4mini}, and DeepResearch ~\cite{openai2025deepresearch_system_card},
as well as leading open-source methods including ASearcher ~\cite{gao2025turnsunlockinglonghorizonagentic}, DeepDive ~\cite{lu2025deepdiveadvancingdeepsearch},
MiroThinker ~\cite{miromindteam2026mirothinkerpushingperformanceboundaries}, InfoAgent ~\cite{zhang2025infoagent}, Kimi-K2 ~\cite{kimiteam2026kimik2openagentic}, WebExplorer ~\cite{liu2025webexplorer}, WebDancer ~\cite{wu2026webdancer}, WebSailor ~\cite{li2025websailornavigatingsuperhumanreasoning}, WebShaper ~\cite{tao2026webshaper}, WebLeaper~\cite{tao2025webleaper}, and C-GSPO ~\cite{zhang2026chainingevidencerobustreinforcement}.

\paragraph{Training Settings and Hyperparameter Details}
Unless otherwise specified, all models are trained based on Qwen3-VL-8B-Thinking and Qwen3-VL-30B-A3B-Thinking.
During rollout generation, we use a sampling temperature of 0.6 and top-p nucleus sampling with p = 0.95. We set the fixed weighting factor to 0.3 and the top-p evidence selection ratio to 10\%.
\subsection{Compute Budget}
All models were trained for 50 steps on NVIDIA H800 GPUs.
The computational cost for the 8B model was approximately 443 GPU-hours.
For the 30B-A3B model, the computational cost increased to approximately 534 GPU-hours.
\subsection{Tools Description}
\paragraph{RAG Text Tool}
To align with other models, facilitate distillation, and provide a robust cold start, we implement an intelligent retrieval and reranking module, \textbf{RAGTool}. The module performs content acquisition, text normalization/chunking, and relevance scoring in a unified workflow. For content acquisition, we employ a high-throughput crawler based on the \texttt{httpx} asynchronous framework, supporting up to \textbf{32 concurrent} requests to efficiently gather candidate pages. For access-restricted or unstable webpages (e.g., dynamic rendering, interstitials, or anti-bot behaviors), we incorporate the JINA API as a fallback mechanism to improve the robustness and coverage of raw text retrieval while maintaining a consistent downstream interface. Retrieved HTML is then cleaned and normalized (removing boilerplate and redundant markup) and segmented under a token-budget constraint to fit embedding/reranking context limitations. Concretely, the base chunk size is set to \textbf{256 tokens}, and a lightweight \texttt{naive\_merge} strategy merges under-length segments with adjacent neighbors to reduce semantic fragmentation at chunk boundaries. For relevance estimation, we adopt \textbf{BAAI/bge-reranker-v2-m3}, which provides strong semantic matching and cross-lingual retrieval capabilities, to compute query--chunk relevance scores. Candidate chunks are ranked in descending order, and only the \textbf{Top-$K$} high-confidence segments are retained (\textbf{$K=10$}). Finally, the concatenated evidence is dynamically truncated to a maximum context budget (default \textbf{2048 tokens}) so that limited downstream context is dominated by high-salience signals, thereby improving the information density for subsequent reasoning.

\paragraph{Full-Text Tool}
To serve as a direct text-based baseline for our text and snapshot comparative experiments, we introduce a straightforward HTML parsing module that bypasses semantic reranking. Following the identical high-throughput content acquisition pipeline (leveraging asynchronous \texttt{httpx} with the JINA API fallback), this tool directly processes the retrieved HTML to extract the complete raw text. After stripping standard boilerplate and redundant structural markup, the entire textual content is retained in its original linear order. To ensure a fair comparison with the visual snapshot modality regarding token consumption, the extracted full text is simply truncated to a maximum token limit that precisely matches the snapshot's context window size. This naive extraction and truncation mechanism deliberately preserves the original flow of the webpage, allowing us to explicitly evaluate the impact of context overflow and token efficiency when comparing uncompressed text representations against multi-modal visual snapshots.
\paragraph{Snapshot Tool}
To mitigate resolution constraints and context loss when vision--language models process long webpages, we develop a visual perception component, \textbf{SnapshotTool}, that integrates high-fidelity rendering, adaptive slicing, and visual optimization within a single pipeline. The system is implemented with Playwright and maintains a \textbf{browser pool of 8 Chromium instances}, supporting up to \textbf{32 concurrent} rendering requests for scalable batch processing. During rendering, an auto-scrolling procedure is applied to trigger lazy-loaded elements and ensure comprehensive visual capture, while a lightweight denoising routine suppresses non-content overlays (e.g., cookie banners, modal dialogs, and popups) through targeted DOM manipulation and style injection to provide clean visual inputs. For long pages, we enforce a hard maximum rendering height of \textbf{20{,}000 pixels}, which covers the primary information region of most websites while bounding worst-case computation. Within this range, a sliding-window strategy produces a sequence of serialized screenshot slices with slice height \textbf{4{,}480 pixels} and vertical overlap \textbf{112 pixels}, reducing boundary truncation artifacts (e.g., text lines, figures, and tables split across slice edges) and preserving visual semantic continuity across slices. To reduce downstream VLM input overhead without sacrificing readability, each slice is uniformly downsampled by a \textbf{0.7$\times$} factor using \textbf{LANCZOS} interpolation. When local rendering fails, the module falls back to the \textbf{Jina API} and applies the same slicing and compression logic, ensuring consistent output structure and comparable visual characteristics across rendering pathways.

\subsection{Data Curation}
\paragraph{Supervised Finetuning Data.}

To equip the model with foundational web-interaction skills and establish a stable initialization for reinforcement learning, we construct a hybrid supervised fine-tuning (SFT) dataset consisting entirely of high-difficulty, multi-step reasoning problems. This complex dataset comprises two complementary subsets distilled from different observation modalities. 

The first subset leverages an automated framework that integrates hierarchical knowledge extraction on Wikipedia with adaptive web exploration to synthesize a pool of sophisticated QA pairs. Trajectories for these complex problems are distilled using DeepSeek under a \textit{text-based} fetching modality. To ensure a fair comparison regarding token budgets, we apply a goal-guided reranking strategy to extract the most relevant textual chunks, intentionally establishing a remarkably strong text baseline. 

The second subset focuses on visual web navigation for similarly challenging tasks, utilizing diverse, high-difficulty information-seeking questions sampled from ASearcher \cite{gao2025turnsunlockinglonghorizonagentic}. For these queries, we distill trajectories using GLM 4.6 under a \textit{snapshot-based} fetching modality, which returns webpage snapshots to explicitly preserve layout and structural cues. By combining these two distillation sources, the SFT phase trains the model to tackle complex web-search mechanics while becoming fluent in both highly-optimized textual and visual web representations.

\paragraph{Reinforcement Learning Data.}

To drive policy optimization during the RL phase, the agent requires challenging environments that demand deep, autonomous exploration. Rather than relying on generic queries, we curate a specialized dataset of 0.4k high-difficulty, multi-layered questions. These queries are explicitly selected from the complex problem pool generated by our aforementioned automated synthesis framework. By utilizing these carefully constructed long-tail questions as the initial prompts, we encourage the RL agent to engage in multi-turn exploration and complex evidence alignment without relying on step-by-step expert demonstrations.
\begin{table}[ht]
\centering
\footnotesize
\caption{Answer accuracy (\%) across varying trajectory lengths (interaction steps). The evaluation is conducted under different evidence filtering thresholds ($p \in \{10, 15, 20\}$). The best results within each threshold setting are highlighted in bold.}
\label{tab:length_controlled_sufficiency}
\renewcommand{\arraystretch}{1.1}
\setlength{\tabcolsep}{5pt}
% 关键修改：将 \linewidth 改为 0.85\linewidth，强行缩小整个表格的比例
\resizebox{0.75\linewidth}{!}{
\begin{tabular}{c l ccccc c}
\toprule
\multirow{2}{*}{\textbf{Threshold}} & \multirow{2}{*}{\textbf{Method}} & \multicolumn{5}{c}{\textbf{Trajectory Length (Steps)}} & \multirow{2}{*}{\textbf{Avg.}} \\
\cmidrule(lr){3-7}
& & \textbf{0--19} & \textbf{20--39} & \textbf{40--59} & \textbf{60--79} & \textbf{80--99} & \\
\midrule

\rowcolor{bg_green}
\multirow{3}{*}{\textbf{$p=10\%$}} 
& Top-$10\%$    & \textbf{69.0} & \textbf{69.0} & \textbf{73.0} & \textbf{67.7} & \textbf{75.0} & \textbf{70.7} \\
& Random-$10\%$ & 67.0 & 55.0 & 52.0 & 55.9 & 45.8 & 55.1 \\

& Tail-$10\%$   & 66.0 & 52.0 & 33.0 & 45.6 & 40.0 & 47.3 \\
\midrule

\rowcolor{bg_green}
\multirow{3}{*}{\textbf{$p=15\%$}} 
& Top-$15\%$    & \textbf{80.0} & \textbf{68.0} & \textbf{68.0} & \textbf{72.1} & \textbf{62.5} & \textbf{70.1} \\
& Random-$15\%$ & 73.0 & 57.0 & 60.0 & 54.4 & 50.0 & 58.9 \\

& Tail-$15\%$   & 61.0 & 50.0 & 45.0 & 45.6 & 45.2 & 49.3 \\
\midrule

\rowcolor{bg_green}
\multirow{3}{*}{\textbf{$p=20\%$}} 
& Top-$20\%$    & \textbf{75.0} & \textbf{69.0} & \textbf{61.0}& \textbf{77.9} & \textbf{79.2} & \textbf{72.2} \\
& Random-$20\%$ & 71.0 & 61.0 & 60.0 & 60.3 & 50.0 & 60.7 \\

& Tail-$20\%$   & 67.0 & 51.0 & 45.0 & 52.9 & 54.2 & 54.0 \\
\bottomrule
\end{tabular}
}
\end{table}

\begin{table*}[hb]
\centering
\footnotesize
\caption{Results on BrowseComp, GAIA, Xbench-DS, Seal-0. The results are evaluated with LLM-as-Judge. For baselines, we run the evaluation for 4 seeds and report \textbf{Pass@4}, which reflects the model's upper bound. Results marked with $^{*}$  are taken from existing studies.}
\begin{tabular}{llccccc}
\toprule
\textbf{Method} & \textbf{Fetch Modality} & \textbf{BrowseComp} & \textbf{GAIA} & \textbf{Xbench-DS} & \textbf{Seal-0}  \\

\midrule
% \rowcolor{lightyellowbg}
% \multicolumn{6}{c}{\textbf{Pass@4}} \\
QwQ-32B Direct Gen.$^{*}$   & Text & --   & 31.1 & 23.0   & --     \\
Search-o1 (QwQ)$^{*}$  & Text & -- & 67.0 & 65.0 & -- \\
Search-R1-32B$^{*}$  & Text & -- & 43.7 & 37.0 & --  \\
WebThinker-QwQ$^{*}$ & Text & -- & 57.3 & 52.0 & --  \\
Simple DS-QwQ$^{*}$ & Text & -- & 64.1 & 61.0 & --  \\
WebDancer-QwQ$^{*}$ & Text & -- & 61.2 & 68.0 & --  \\
ASearcher-Web-QwQ-v2$^{*}$ & Text & -- & 70.1 & 68.0 & -- \\
\rowcolor{highlightblue}
Qwen3-VL-8B-ICA (Ours)    & Snapshot & \uline{39.9} & \uline{86.4} & \uline{87.0} & \uline{49.3}  \\
\rowcolor{highlightblue}
Qwen3-VL-30B-A3B-ICA (Ours)   & Snapshot & \textbf{45.0} & \textbf{89.3}& \textbf{89.0} & \textbf{52.3} \\
\bottomrule
\end{tabular}
\label{tab:information_seeking}
\end{table*}

\begin{algorithm}[ht]
\caption{ICA-GSPO}
\label{alg:decoupled_rl}
\begin{algorithmic}[1]

\REQUIRE Policy $\pi_\theta$, question $q$, high-utility evidence sets $\{\mathcal{T}^c\}$, maximum context length $L_{\max}$
\ENSURE Updated policy parameters $\theta$

\STATE Sample a group of trajectories $\{\tau_i\}_{i=1}^{N} \sim \pi_{\theta_{\text{old}}}(\cdot \mid q)$
\STATE Evaluate final answers to obtain correctness outcomes $\{R_i\}_{i=1}^{N} \in \{0, 1\}$

\FOR{each trajectory $\tau_i$}
\STATE \textit{1. Compute budget-aware shaped reward for the final answer}
    \IF{$R_i = 1$}
        \STATE $\tilde{R}_i \gets 1.0$
    \ELSE
        \STATE $\tilde{R}_i \gets -1.0 + \sqrt{|\tau_i| / L_{\max}}$
    \ENDIF
    
\STATE \textit{2. Compute binary rewards for intermediate tool calls}
    \FOR{each tool-invoking turn $t$ with tool type $c_t^i$}
        \STATE Extract acquired evidence/URL set $\mathcal{E}_i$
        \STATE $r_t^i \gets \mathbb{I}\bigl[ \mathcal{E}_i \cap \mathcal{T}^{c_t^i} \neq \emptyset \bigr]$
    \ENDFOR
\ENDFOR

\STATE \textit{3. Compute group-level normalization statistics}
\STATE Compute $(\bar{R}, \sigma_R)$ over final shaped rewards $\{\tilde{R}_i\}_{i=1}^{N}$
\STATE Compute $(\bar{r}, \sigma^r)$ over $\{r_t^i\}$ independently for each tool type $c \in \{\textsc{search}, \textsc{fetch}\}$

\FOR{each trajectory $\tau_i$}
    \STATE \textit{\# 4. Assign turn-level decoupled advantages}
    \FOR{each turn $t$ in $\tau_i$}
        \IF{$t$ is a tool-call turn of type $c_t^i$}
            \STATE $A_t^i \gets \frac{r_t^i - \bar{r}}{\sigma_{r} + \epsilon}$
        \ELSIF{$t$ is the final answer turn}
            \STATE $A_t^i \gets \frac{\tilde{R}_i - \bar{R}}{\sigma_R + \epsilon}$
        \ENDIF
    \ENDFOR
\ENDFOR

\STATE Update $\theta$ using policy gradient/GSPO surrogate loss with advantages $A_t^i$

\end{algorithmic}
\end{algorithm}
To facilitate reproducibility, we provide the pseudocode of our ICA in Algorithm \ref{alg:decoupled_rl}.
\section{Additional Experimental Results}
Table \ref{tab:information_seeking} summarizes results on BrowseComp, GAIA, XBench-DS, and Seal-0 under an LLM-as-Judge evaluation protocol. These benchmarks are designed to stress web agents' ability to browse real pages, follow long-horizon reasoning chains, and ground answers on external evidence rather than relying on the LLM's internal knowledge. Among the text-fetching baselines, ASearcher-Web-QwQ-v2 is the strongest on GAIA (70.1) and XBench-DS (68.0). In contrast, our snapshot-based ICA consistently delivers the best performance across all reported datasets: ICA-8B already reaches 86.4 on GAIA and 87.0 on XBench-DS, and scaling to ICA-30B-A3B further boosts results to 89.3 and 89.0, surpassing the best text baseline by +19.2 and +21.0 points, respectively. Moreover, ICA-30B-A3B achieves 45.0 on BrowseComp and 52.3 on Seal-0, demonstrating robust gains on challenging browsing-centric tasks. For baselines, we run the official implementations with four random seeds and report Pass@4, which reflects the model's upper-bound performance.
\section{Judge Prompt}
\begin{tcolorbox}[
    breakable,                    % 开启跨页支持 (关键修改)
    colback=gray!15,              % 框内背景颜色 (浅灰)
    colframe=gray!75!black,       % 边框和标题背景颜色 (深灰)
    title=Judge prompt,           % 标题替换为 Judge prompt
    fonttitle=\bfseries,          % 保留标题栏原本的加粗以维持美观，但正文完全统一
    coltitle=white,               % 标题文字颜色 (白色)
    arc=3mm,                      % 圆角弧度
    boxrule=1.5pt,                % 边框粗细
    left=5mm,                     % 左侧内边距
    right=5mm,                    % 右侧内边距
    top=4mm,                      % 顶部内边距
    bottom=4mm,                   % 底部内边距
    toptitle=2mm,                 % 标题顶部内边距
    bottomtitle=2mm               % 标题底部内边距
]

You grade whether a model's answer matches the standard answer for a benchmark question.

\vspace{0.5em}
Inputs:
\begin{itemize}
    \item {[Question]}: the task.
    \item {[Standard Answer]}: the reference (treated as correct).
    \item {[Model\_answer]}: the model's full or post-processed output (may include reasoning, citations, or how it found the answer).
\end{itemize}

\vspace{0.5em}
Your job: decide if the substantive content of [Model\_answer] matches [Standard Answer] for what the question asks.

\vspace{0.5em}
Output 1 (correct) when:
\begin{itemize}
    \item The model's final factual claim matches the standard answer, even if phrased differently, in different order, or with extra units/formatting (e.g. 0.1777 m$^3$ vs 0.1777, "Annie Levin" vs "**Annie Levin**").
    \item Extra reasoning, search steps, quotes, footnote markers like [\textasciicircum{}0\textasciicircum{}], or URLs are OK if the resolved answer still matches.
\end{itemize}

\vspace{0.5em}
Output 0 (incorrect) when:
\begin{itemize}
    \item The model gives a different number/name/date/entity, contradicts the standard answer, or is off-topic.
    \item The model clearly refuses or says it cannot answer.
    \item The model only restates the question or only copies the standard answer without committing to an answer (no substantive solution).
\end{itemize}

\vspace{0.5em}
Important: Describing that information was found via search, browsing, or tools does NOT by itself make the answer wrong. Grade the correctness of the answer content, not the presence of tool narration.

\vspace{0.5em}
If they are consistent, Judgement is 1; if they are different, Judgement is 0.

\vspace{0.5em}
Just output Judgement and don't output anything else.

\end{tcolorbox}
\end{document}